\documentclass{article}

\usepackage{arxiv}

\usepackage[utf8]{inputenc} % allow utf-8 input
\usepackage[T1]{fontenc}    % use 8-bit T1 fonts
\usepackage{hyperref}       % hyperlinks
\usepackage{url}            % simple URL typesetting
\usepackage{booktabs}       % professional-quality tables
\usepackage{amsfonts}       % blackboard math symbols
\usepackage{nicefrac}       % compact symbols for 1/2, etc.
\usepackage{microtype}      % microtypography
\usepackage{lipsum}

\usepackage{cite}
\usepackage{amsmath,amssymb,amsfonts}
\usepackage{algorithmic}
\usepackage{graphicx}
\usepackage{textcomp}
\usepackage{soul}
\usepackage{xcolor,colortbl}
\usepackage{float}
\usepackage{longtable}
\usepackage{hyperref}
\hypersetup{colorlinks,allcolors=black}
\def\BibTeX{{\rm B\kern-.05em{\sc i\kern-.025em b}\kern-.08em
    T\kern-.1667em\lower.7ex\hbox{E}\kern-.125emX}}

\title{A Survey on Churn Analysis}

\author{
  Jaehuyn Ahn \\
  Gyeonggi-do, Republic of Korea \\
  \texttt{jaehyunahn@sogang.ac.kr} \\
}

\begin{document}
\maketitle

\section{Introduction}
\label{sec:intro}

The term customer churn is commonly used to describe the propensity of customers who cease doing businesses with a company in a given time or contract . Traditionally, studies on customer churn started from Customer Relation Management (CRM) . 
It is crucial to prevent customer churn when operating services. In the past, the efficiency of customer acquisition relative to the number of churns was good. However, as the market saturated because of the globalization of services and fierce competition, customer acquisition costs rose rapidly . 

Reinartz, Werner, Jacquelyn S. Thomas, and Viswanathan Kumar. (2005) have shown that, for long-term business operations, putting efforts to increase the retention rate of all customers in terms of CRM is less efficient than putting efforts on a small number of targeted customer acquisition activities . Similarly, Sasser, W. Earl. (1990) have suggested that retained customers generally return higher margins than randomly targeting new customers . Additionally, Mozer, Michael C., et al. (2000) have proposed that, in terms of net return on investment, marketing campaigns for retaining existing customers are more efficient than putting efforts to attract new customers . Reichheld et al. (1996) have shown that a 5 percent increase in customer retention rate achieved 35 percent and 95 percent increases in net present value of customers for a software company and an advertising agency, respectively . As such, churn prediction can be used as a method to increase the retention rate of loyal customers and ultimately increase the value of the company. 

Studies on customer churn have been proposed in various service fields. These studies on the churn analysis attempted to identify or predict in advance the likelihood that customers will churn using various indicators. The customer churn rate  is a typical customer churn analysis indicator. This refers to the ratio of subscribers who cancel a service to the total number of subscribers during a specific period . The churn rate is the most widely used indicator for calculating the service retention period of subscribers in most service fields. Because of its importance and intuition, churn has been introduced in various service fields and developed to suit the characteristics of each field. Consequently, the research on the analysis of customer churn was fragmented according to each research field, thus the measurement criteria are all different. Currently, this is causing many problems. In the industry, communication costs arising from different churn criteria between service personnel in the process of fusing heterogeneous services (e.g., vehicle sharing service/insurance, online music service/department store) have been sharply increasing. Furthermore, since research on churn is simultaneously associated with two fields of engineering and business administration, it is not easy for researchers to describe two separate specialized fields on a single paper or to understand them.  

In the past, customer churn of early days was used to define the customer’s status in the CRM. The CRM is a business management method that first emerged as a way of increasing the efficiency in areas of retail, marketing, sales, customer service, and supply-chain, and increasing efficiency and the customer value functions of the organization . Since then, in the architectural point of view, the CRM has evolved and become divided into operational CRM and analytical CRM. The analytical CRM is focused on developing databases and resources containing customer characteristics and attitudes . The analytical CRM has been initially used for creating appropriate marketing strategies using customer status and customer behavior data, and particularly, it has been used to fulfil the individual and unique needs of customers . From this point on, IT and knowledge management related technologies have been utilized, and companies started applying dedicated technologies for acquisition, retention, churn, and selection of customers , and ever since the technologies of IT field became implemented in the CRM, various companies began to use such technologies in business areas including data warehouse, website, telecommunication, and banking . As described earlier, with studies on CRM claiming that increasing the retention rate of small number of existing customers is more efficient than acquiring new customers, churn analysis has become one of the important personalized customer management techniques . There were survey papers that collected and summarized churn analysis techniques in the telecommunications field . However, these studies are limited to the telecommunications field, and the log data used for the churn analysis do not include time series features, retention and survival, and KPI (Key Performance Indicator) features. There were also papers applied to services using various deep learning model-based churn analysis techniques in terms of computer science . However, these studies are limited to the deep learning algorithm, and lack underlying models and parameter description. There are also a few survey papers on churn, yet they do not cover the latest deep learning techniques but cover only churn in specific industrial fields . The trend of building churn prediction models is changing, and performance is rapidly improving. However, because of fragmented previous studies, there are many difficulties for researchers to launch new research on churn. In order to address these issues, this survey paper describes the differences in the definition of churn prediction algorithms in the fields of business administration, marketing, IT, telecommunications, newspaper publishing, insurance, and psychology, and compares differences in churn loss and feature engineering. In addition, I classify and explain the cases of churn prediction models based on this. Our study provides classification information for more detailed technologies on churn in a wider range than previous survey papers. Our research can reduce confusion about the churn criteria that are being fragmented and utilized across multiple industry/academic fields, and can be of a practical help in applying them to prediction models. In particular, this paper presents a deep learning model among machine learning techniques designed to solve non-contractual customer churns, which have recently appeared with the advancement of industries. The structure of this paper is as follows.

Churn has been defined in various ways in multiple industries. In this chapter, I describe two typical types.  Typical papers with different criteria for defining churn are summarized. In general, the dictionary definition of churn is known as the prolonged period of inactivity . However, the criteria for \lq inactivity\rq and \lq prolonged\rq are different according to each research field. Such inconsistency is frequently found due to more services of modern days adopting loose subscription terms because of competition. In the past, customer churn had occurred explicitly through contractual cancellations, however, in the modern services including Internet and retail services, frequent customer churns occur due to the low customers’ investment costs . These non-contractual customer churns occur due to low switching cost for changing the service . Thus, I was divide the criteria of churn into contractual churn and non-contractual churn. Descriptions of each churn is as follows. 

The first criterion is contractual churn. Contractual churn refers to churn that a customer does not extend the contract even when the contract renewal date is reached . This churn means that a customer loses interest in the relevant service area and changes his/her position to a state where re-entry is no longer possible. It is usually present in churn problems occurring when customers close their banking accounts or when switching their carrier operator from one service to another. In addition, contractual churn is frequently found in a flat-rate service such as music and movie streaming services.

The second criterion is non-contractual churn. 
% 답변 16
% 일반적으로 non-contractual situation에서 customers는 제약 없이 서비스/계약을 이탈할 수 있다.
In general, in a non-contractual situation, customers can leave the service/contract without time constraints.
In the service operating perspective, a criterion for churn is first constructed, then a customer that meets such criterion is categorized as the churn customer. To conduct this, the customer’s behavioral changed date is counted . When this inactivity or behavioral changed period exceeds the threshold, the customer is regarded as a churn customer. During this process, the period that is set as the threshold of the inactivity date is called the time window . The defining of non-contractual churn has made it possible to infer the probability of the customers who are likely to churn within the certain period.
The time window method is frequently used when analyzing activity logs these days in a non-contractual situation. When a customer does not use a service for a certain period of time, this method regarding customer as churned. Internet services do not usually delete accounts. Therefore, the Internet service interprets the log-in as prolonged, that is, the retention of the service, and interprets unconnected access for a certain period of time as churn.

Churn analysis is usually performed to improve business outcomes. Therefore, in most churn prediction problems, the churn period is defined as a section that can restore customers' trust. If the time period during which a customer completely churns is selected as a time window, the period for churn definition exponentially increases and it does not provide any gain in terms of business as changing the will of the customers who want to churn is deemed impossible . The contractual mentioned above are close to customers’ complete churn from a service. Therefore, these days majority of the log-based churn prediction problems use the probabilistic method to determine whether customers are churning or not and to give customers incentive to reuse their service.

As described above, there are two customer churn types, which are contractual churn and non-contractual churn. Additionally, there are three churn observation criteria as follows: monthly, daily, and binary. The monthly and daily churn observations are related to the cycle in which the customer’s status is updated in the database. The binary churn observation is acquired by manipulating this database. In general, binary churn is determined by the existence of contract in the contractual settings. In the non-contractual settings, the company defines the customer inactivity features, and when a customer meets the inactivity or disloyal customer feature, the customer is regarded as binary churn . The reason for having multiple ways of defining customer churn is to periodically monitor the customers’ status changes. And through such observation, the expected net business value can be increased by predicting customer churn rates and providing possible churn customers with incentives to retain them from leaving .

\section{Churn Analysis in Various Business Field}
\label{sec:ch3}

The majority of the early studies on churn were conducted from a management perspective, especially CRM (Customer Relation Management) . CRM churn covers all churn problems that can occur in the process of customer identification, customer attraction, customer retention, and customer development. Modern churn prediction problems are mainly analyzed using log data. A log is trace data that remains when using Internet services. Therefore, the churn prediction models implemented using log data can be used for Internet services in various industries. There are 12 business fields that performed churn prediction.

The telecommunications industry accounts for the majority of previous studies on churn. Telecommunications services have high customer stickiness despite high customer acquisition costs. Therefore, if customer churn is prevented and appropriate incentives are provided, it is of great help in maintaining sales .

The financial and insurance industries also predict customer churn. Zhang, Rong, et al. (2017) stressed the need to build churn prediction models and prevent churn, referring to high customer acquisition costs and high customer values in the insurance industry . Chiang, Ding-An, et al. (2003) mentioned that customer values were high in the online financial market, and created a churn scenario according to the financial product selection and customers' financial product selection sequence using the Apriori algorithm . Larivière, Bart, and Dirk Van den Poel. (2004), based on the assumption that the customer group was different according to the financial product attribute, demonstrated that the likelihood of churn differed depending on the tendency of customers who selected financial products by measuring the survival time for each product . Zopounidis, Constantin, Maria Mavri, and George Ioannou. (2008) measured the switching rate of financial products, and the survival period of customers for each product to discover attractive products . Here, as the survival period is short, churn occurs more frequently, which is used as an indicator to measure the need to supplement financial products. Glady, Nicolas, Bart Baesens, and Christophe Croux. (2009) measured the customer lifetime values and the decrease in expected earnings over time as an indicator corresponding to customer loyalty . During this process, machine learning was used to calculate the churn rate which was used to estimate the customer lifetime values.

Later on, studies on churn have been actively conducted in the gaming field as in the telecommunication field. These services have a fast cycle of customer inflow and churn because of mass competition. However, if a single service is run for a long time, the service competition intensifies and the Customer Acquisition Cost (CAC) tends to increase . As the CAC gets larger, the technology to predict and prevent churn becomes more crucial. Viljanen, Markus, et al. (2016) applied the survival analysis to mobile games and calculated the churn rate, similar to the churn prediction of financial services . The game sector actively uses machine learning techniques when conducting research on churn because of the large volume of log data . Milošević, Miloš, Nenad Živić, and Igor Andjelković. (2017) created a model predicting churn in the study on game churn, gave churn prevention incentives by finding out and dividing probable churn customers into A/B groups, and demonstrated actual effects statistically . Runge, Julian, et al. (2014) conducted a similar study, and revealed that existing customers with a high possibility of churn had a higher marketing response rate when compared to general marketing targets .

Furthermore, the music streaming service field even held a competition to build a prediction model, and research on churn was also conducted in the Internet service and newspaper subscription fields. The newspaper subscription and music streaming service offer fixed-rate services, and customer churn is consistent with the contract renewal period. On the other hand, because the Internet service goes into an inactive state as customers wish, contract renewal takes place nearly-real-time. Research on churn prediction was also conducted in online dating, online commerce, Q\&A services, and social network-based services 

There were some studies which approached customer churn from a psychological perspective. Borbora, Zoheb, et al. (2011) analyzed that customers churned when their motivation to use games changed by combining the motivation theory with customers using MMO RPG games . Yee, Nick. (2016) surveyed approximately 250,000 gamers, and showed that customers' attitudes toward games were clustered by country, race, and age .

In the marketing field, Glady, Nicolas, Bart Baesens, and Christophe Croux. (2009) used the features from a marketing perspective such as RFM (Recency, Frequency and Monetary) and CLV (Customer Life time Value) for churn prediction .

Studies on churn prediction were conducted in the human resources and energy fields although they were minority.  Saradhi, V. Vijaya, and Girish Keshav Palshikar. (2011) conducted research on churn to reduce retraining costs when employees churned and to prove employee value in the human resources field . Moeyersoms, Julie, and David Martens. (2015) estimated whether customers would churn to another energy supplier based on energy data and socio-demographic data provided to customers .

% 답변 2
\section{Conclusions}
\label{sec:ch8}

In this study, I compared the churn prediction analysis techniques using log data. Churn analysis is used in the fields of Internet services and games, insurance, and management. Research on churn prediction usually begins to improve business outcomes. Therefore, the time window is used to select potential churning customers rather than measuring a customer's complete churn. Loss costs for customer churn are calculated by CAC or CLV. In the past, when predicting customer churn, researchers used survival analysis or time series analysis using statistics, graph theory, and traditional machine learning algorithms. Churn prediction analysis using deep learning algorithms has recently emerged. Deep learning algorithms have been found to outperform other algorithms. 
% 그 이유(딥러닝이 과거 다른 알고리즘보다 퍼포먼스가 좋은 이유)는 로그 데이터가 전산을 통해 대량으로 수집되기 때문이며, 이탈 예측 모델은 이렇게 주어진 고객의 모든 로그를 종합하여 이탈을 예측해야 하기 때문일 것이다. 본 논문 VI. Churn prediction models에 소개한 논문 중 딥러닝을 사용하여 이탈을 예측한 논문들의 경우, 고객 로그 데이터의 데이터 timestamp가 seconds 단위이거나 데이터 총량이 매우 컸다. 이럴 경우 로그를 가공하는 feature engineering 기법이 모델 성능 향상에 매우 큰 영향을 미친다. 딥러닝 모델은 위에서 소개했던 다른 모델링 기법들과는 다르게 time series features를 embedding 해서 high dimension and sparse한 log 데이타를 low dimension and dense한 features로 바꿀 수 있다. 따라서 로그의 timestamp가 상세 하고, 수집되는 관찰 timestamp가 많을 때, 이 데이터를 딥러닝 알고리즘의 latent features 생성 기법에 사용하면 기존 churn prediction models보다 나은 성능을 보일 것을 기대한다.

This is because as the log data in these days is collected for a longer period and deep learning algorithm get an advantage to catch customers' latent status compared with older algorithms.
In other words, the reason deep learning algorithms are receiving spotlight today is due to the vast amount of data used in modern churn predictions, and its ability to capture minute changes. As mentioned earlier in the text, traditional churn prediction algorithms including statistics methods are still actively used today. This is due to variations in which churn prediction model has the best performance depending on the data format. Churn prediction models using deep learning is a new solution with a good structure for predicting modern churn datasets. Therefore, to solve the problem at hand, readers will need to understand the format of the churn dataset and apply a suitable algorithm to solve the churn prediction problem.
% 다시말해 딥러닝 알고리즘이 현대 churn prediction algorithm으로 각광받는 이유는 오늘날 이탈 예측에 사용되는 데이터의 양이 과거에 비해 방대하고, 변화를 민감하게 포착하기 적합한 구조이기 때문이다. 앞서 본문에서 언급했듯 statistics method를 포함한 과거 churn prediction algorithm은 오늘날에도 활발히 사용된다. 그 이유는 데이터의 형식에 따라 최상의 성능을 내는 churn prediction model의 성능이 각자 다르기 때문이다. deep learning을 사용한 churn prediction model은 현대의 이탈 데이터세트를 예측하기 좋은 구조를 가진 새로운 해결책이다. 그러므로 독자들은 해결해야 하는 문제의 churn dataset의 형식을 이해하고, 각자의 데이터에 적합한 알고리즘을 사용하여 churn prediction problem을 해결해야 할 것이다.
% ----

Furthermore, I also outlined a performance evaluation method for comparing the various churn prediction algorithms used from the past to the present. Most churn prediction models are related to customer relation management. For example, there may be performance differences depending on whether the churn prediction model is robust against false positives or false negatives. According to the research of this paper, many articles use AUC as a performance measurement method aside from standard precision. In general, as there are fewer churn customers than non-churn customers, a performance specific method focused on churn customers will be needed. The ROC curve is a graph of the rate at which the model correctly predicts churn customers and the rate at which residual customers are predicted to be churn customers. Therefore, it is a performance measurement method that focuses on the prediction of churn customers.
% 또한 우리는 과거에서부터 지금까지 사용된 다양한 이탈 예측 알고리즘을 비교할 수 있는 performance evaluation 방법도 정리했다. 이탈 예측 모델은 대부분 Customer Relation Management와 연결되어 있다. 예를 들어 이탈 예측 모델이 False Positive 혹은 False Negatie에 강한지에 따라 퍼포먼스의 차이를 보일 수 있다. 본 논문이 조사한 바에 의하면 다수의 논문들이 일반적인 Precision 또는 F1-Score 외에도 AUC를 퍼포먼스 측정 방식으로 사용한다. 일반적으로 이탈 고객은 비이탈 고객에 비해 소수이기 때문에 이탈자에 focus가 된 퍼포먼스 특정 방식이 필요할 것이다. ROC 커브는 모델이 이탈자를 제대로 예측하는 비율과 잔존 고객을 이탈자라고 예측하는 비율에 대한 그래프이다. 따라서 이탈자 예측에 초점이 맞춰져 있는 퍼포먼스 측정 방법이다.
In this study, I comprehensively compared the churn prediction problems. This paper helps to find a method that meets the needs of researchers among various churn prediction algorithms. Furthermore, this paper is expected to be used to improve services and build better churn analysis models.

% Appendix 부분: https://tex.stackexchange.com/questions/347621/ieeetran-problem-in-appendix
% Appendix 위치: IEEE Writiing Principles https://ieeeauthorcenter.ieee.org/wp-content/uploads/IEEE_Style_Manual.pdf

% 주석 References

\end{document}